\pdfoutput=1
\documentclass[11pt]{article}

\usepackage{acl}

\usepackage{times}
\usepackage{latexsym}

\usepackage[T1]{fontenc}
\usepackage[utf8]{inputenc}

\usepackage{microtype}

\usepackage{inconsolata}
\usepackage{arydshln}
\usepackage{hyperref}
\usepackage{fontawesome}
\usepackage{graphicx}
\usepackage{enumitem}
\usepackage{multirow}
\usepackage{xcolor}
\definecolor{myblue}{RGB}{0, 59, 113}
\usepackage{colortbl}
\usepackage{booktabs}
\usepackage{xspace}

\newcommand{\redinfobox}{\raisebox{-1pt}{\includegraphics[width=1em]{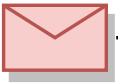}}\xspace}

\newcommand{\grayinfobox}{\raisebox{-1pt}{\includegraphics[width=1em]{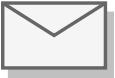}} \xspace}

\newcommand{\greeninfobox}{\raisebox{-1pt}{\includegraphics[width=1em]{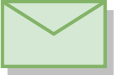}} \xspace}

\newcommand{\lighting}{\raisebox{-1pt}{\includegraphics[height=1em, width=1em]{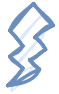}} \xspace}

\title{\includegraphics[width=16pt,height=16pt]{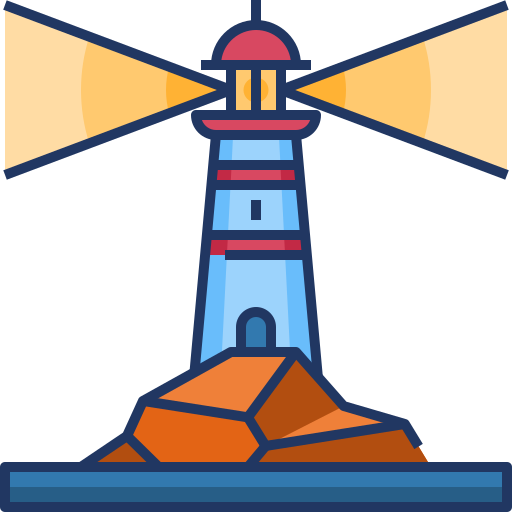} LightHouse: A Survey of AGI Hallucination}

\author{Feng Wang \\
  Soochow University \\
  \texttt{zurichrain403@gmail.com} \\~\\
{{\small \faStar} 
  \texttt{\url{https://github.com/ZurichRain/AGI-Hallucination}}}
  }

\begin{document}
\maketitle

\begin{abstract}
With the development of artificial intelligence, large-scale models have become increasingly intelligent. However, numerous studies indicate that hallucinations within these large models are a bottleneck hindering the development of AI research. In the pursuit of achieving strong artificial intelligence, a significant volume of research effort is being invested in the AGI (Artificial General Intelligence) hallucination research. Previous explorations have been conducted in researching hallucinations within LLMs (Large Language Models). As for multimodal AGI, research on hallucinations is still in an early stage. To further the progress of research in the domain of hallucinatory phenomena, we present a bird's eye view of hallucinations in AGI, summarizing the current work on AGI hallucinations and proposing some directions for future research. We will continuously update recent work on \url{https://github.com/ZurichRain/AGI-Hallucination}.

\end{abstract}

\begin{flushright}
\rightskip=0.8cm\textit{``Once this problem is solved, the path to AGI unfolds. We called it LightHouse for AGI.''} \\
\vspace{.2em}
\rightskip=.8cm---\emph{Anonymous}
\end{flushright}
% \hypersetup{linkcolor=myblue}
% \begin{spacing}{0.8}
% \tableofcontents
% \end{spacing}
\section{Introduction}
The research in deep learning on AGI has witnessed explosive growth, particularly with the advent of LLMs like GPT4 \cite{GPT4}, LLaMA \cite{touvron2023llama}, accelerating the arrival of the AGI era. LLMs have achieved remarkable results in many downstream natural language tasks. Simultaneously, the development of multimodal large models, such as LLaVA \cite{liu2023visual}, has sprung up like mushrooms after rain. There have also been outstanding achievements in the fields of vision, audio, 3D, and agent \cite{lin2023video, Gong2023ListenTA, Hong20233DLLMIT, Szot2023LargeLM}. 

\begin{figure}
    \centering
    \resizebox{.49\textwidth}{!}{
    \includegraphics{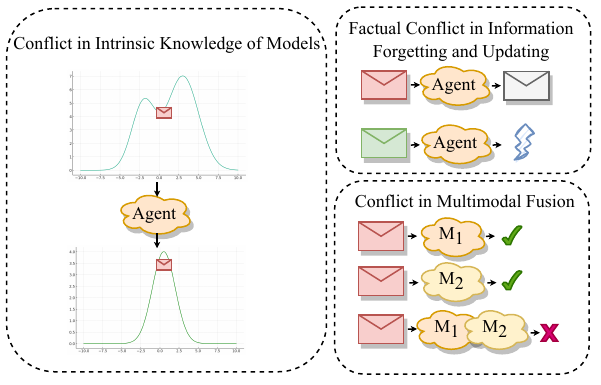}}
    \caption{
     Illustration of Hallucination in AGI. Which classify by three types: 1. Conflict in Intrinsic Knowledge of Models. 2. Factual Conflict in Information Forgetting and Updating. 3. Conflict in Multimodal Fusion. \redinfobox{} is input information, \grayinfobox{} is forgotten information, \greeninfobox{} is new information, and \lighting{} is error output.}
    \label{fig:intro}
\end{figure}

However, despite the astonishing progress in the LLM of AGI, the outputs of the model still do not fully align with human expectations, which called hallucination \cite{Zhang2023SirensSI}. For instance, LLMs occasionally produce factually incorrect answers, LVLMs(Large Vision-Language Models) can sometimes generate responses that object error or something of their own fantasy, video language models also suffer actions inconsistency, and 3D or agent may encounter issues with hallucination across multiple perspectives. This problems hinders AGI's path towards greater intelligence.

To mitigate the hallucinations in models during the AGI era, extensive works are underway. Currently, sparking in LLMs, RLHF \cite{Christiano2017DeepRL, Ouyang2022TrainingLM} (Reinforcement Learning from Human Feedback) is increasingly recognized as a prevalent method for mitigating hallucinatory phenomena in AI models. Enhances the model's responses to more closely align with human preferences through a reward-based RL mechanism. In LVLMs, LURE \cite{Zhou2023AnalyzingAM} used a hallucination revisor to reduce model's output hallucinations. There has a lot works engaged in fine-grained image captions to solve this problem. The 3D-LLM \cite{Hong20233DLLMIT} employs 3D point clouds and additional 3D features as inputs, integrating the pre-trained knowledge of large models to uniformly address a variety of 3D-related tasks which include captioning, dense captioning, 3D question answering, task decomposition, 3D grounding, 3D-assisted dialogue, navigation et al. Additionally, significant efforts in the fields of video, audio, and agent domains have been undertaken to mitigate the hallucination inherent in models. They aslo used fine-grained mutimodel feature to improve model performance. Although hallucinations are not invariably detrimental, in certain instances, they can stimulate a model's creativity. Achieving equilibrium between hallucination and creative model output a significant challenge. Presently, there is an increase in research \cite{Zhang2023UserControlledKF, Yao2023LLMLH} specifically talking about this problem.

Evaluating the hallucination error in AGI is also important area. By assessing LLMs, we gain deeper insights into their capabilities and limitations \cite{huang2023ceval, Chiang2023CanLL, Chan2023ChatEvalTB}. Enhanced evaluation methodologies can significantly improve the interaction between humans and LLMs. This, in turn, has the potential to drive innovative designs and implementations in future interactions. Given the wide-ranging applications of LLMs, it is of utmost importance to ensure their safety and reliability, especially in sectors where safety is paramount, such as financial \cite{Sarmah2023TowardsRH} institutions and healthcare facilities \cite{cai2023medbench}. There also has been a lot of progress in evaluating hallucinations with multimodal large models, including rule-based evaluation \cite{Li2023EvaluatingOH, Lovenia2023NegativeOP}, large model-based evaluation \cite{Wang2023EvaluationAA, Chan2023CLAIREI}, and human-based evaluation \cite{Xu2023LVLMeHubAC, Liu2023ModelsSH}.

To improve clarity in understanding AGI hallucination within the AGI research and community and pave the way for the future research of AGI models, we are strongly motivated to compile this comprehensive survey. The structure of this article can be summarized in several parts as follows. First, we elucidated the concept of AGI Hallucination from three perspectives. Following this, we explore the emergence of AGI Hallucination. Then, we discuss the mitigation for AGI Hallucination within different domains. Finally, we  talk about the evaluatation of AGI Hallucination, and discuss future research directions.

\section{Definition for AGI Hallucination}
Hallucination is a significant problem in AGI, research on hallucinations was predominantly derived from studies on LLMs in earlier stages. Recently, there has been an increasing focus on multimodal studies. In the early stages, researchers possessed an indistinct conceptualization of hallucinations, primarily focusing on the accuracy capabilities of models. Nowadays, we defined hallucinations as \textbf{model outputs that do not align with the contemporary empirical realities of our current  world}. To further investigate hallucinations within AGI, we categorize them into the following three types: 
\begin{itemize}[label=$\circ$]
\item Conflict in Intrinsic Knowledge of Models
\item Factual Conflict in Information Forgetting and Updating
\item Conflict in Multimodal Fusion
\end{itemize}

\subsection{Conflict in Intrinsic Knowledge of Models}
Extensive research \cite{Shen2021TowardsOG} indicates that there is a bias between the output distribution of the models and the distribution of the training data itself. Specifically, the model may produce hallucinations in the output due to learning biases.
The investigation of such hallucinations in AGI is characterized as follows:

\textbf{Language}\ \ \ In LLMs, this type of hallucination often manifests as conflicts between model outputs and prompt, or inconsistencies within the context of the model outputs. As shown in Figure \ref{fig:example}, the task is extracting all spatial relationships from a sentence, yet the model's response lacks many of the spatial relationships present in the input sentence which is a conflict between model outputs and model intrinsic knowledge.

\textbf{Vision-Language}\ \ \ Object Hallucination often occurs in LVLMs. For instance, in Figure \ref{fig:example}, there is a computer next to the water cup in the image, but the model's response suggests that there is nothing beside the water cup. In some cases, there are also object errors. For example, the image shows 'a girl driving a motorcycle,' but the model's answer is 'the girl riding a horse'.

\textbf{Video-Language}\ \ \ \citet{Ullah2022ThinkingHF} shows that there are two kinds of hallucination in the Video-captioning: object and action
hallucination. Object hallucination similar to Vision-Language. While action hallucination is shown in Figure \ref{fig:example}. The caption is "A woman is drawing a paper", and the ground truth is "A woman is folding a paper". 

\textbf{Audio-Language}\ \ \ The model tends to use its language capability to answer the free-form open-ended question instead of conditioning on the audio input. And audio-language models also has a problem with “where is speaking part”. Figure \ref{fig:example} shows an error example that model confuses object recognition at the junction between text and audio.

\textbf{3D-Language}\ \ \ The hallucination of 3D is more complex than that of 2D, as it cannot be fully expressed from a single viewpoint, leading to inconsistencies in understanding across multiple perspectives. For example, in Figure \ref{fig:example}, a 3D point cloud feature is provided along with two questions: "What does it look like from the front?" to which the model responds, "A horse." and "What does it look like from the left side?" to which the model responds, "A dog." . In this instance, the model exhibits inconsistency in its responses across multiple perspectives of the object.

\begin{figure*}
    \centering
    \resizebox{1\textwidth}{!}{
    \includegraphics{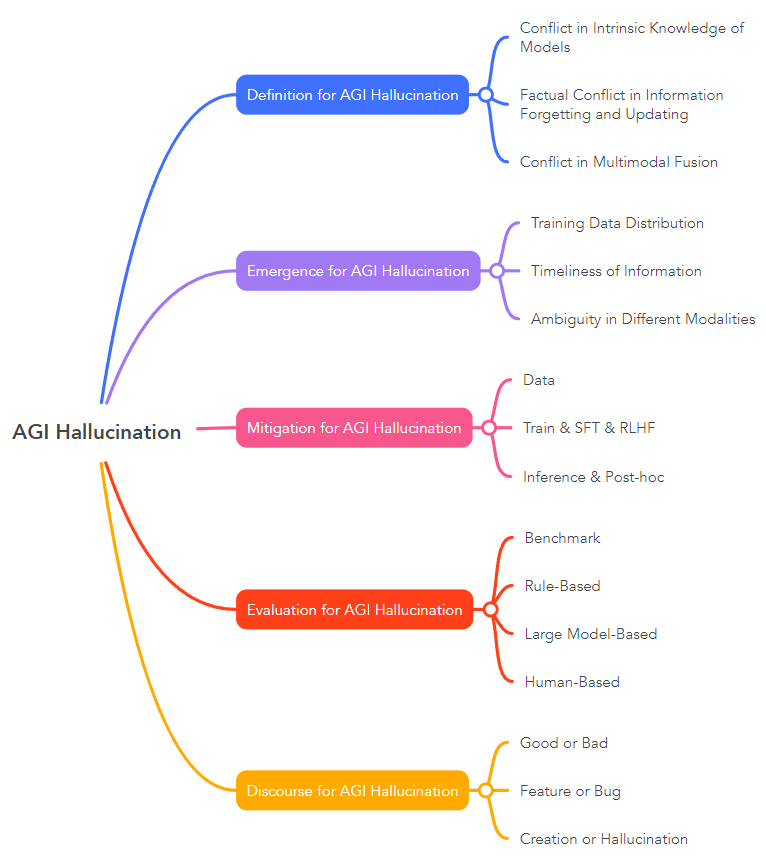}}
    \caption{
    The overview structure of this survey. we have explored the definition of hallucinations in the AGI, examined their causes, evaluated current works to mitigate them, assessed methods for hallucination analysis, engaged in dialogues concerning these phenomena, and contemplated future outlooks in the AGI.}
    \label{fig:overview}
\end{figure*}

\subsection{Factual Conflict in Information Forgetting and Updating}
Factual Conflict primarily arises when models fail to retain previously acquired factual knowledge and are unable to assimilate new information. Previous research on Factual Conflict has mainly focused on language models; recent research shows that the problem of Factual Conflict persists in multimodal contexts. In the following, we will explain the details of Factual Conflict exists in various modalities.

The  hallucinations in the language models is frequently associated with the extent of the model's underlying knowledge base \cite{Augenstein2023FactualityCI, Melz2023EnhancingLI}. They also studied various scenarios in which models produce factual errors. Figure \ref{fig:fact_example} also shows an example of this hallucination. In large image-text models, factual conflicts manifest in two ways. Firstly, the facts contained within the question may be influenced by an object or text in the image. For example in Figure \ref{fig:fact_example}, the text in image is "Barack Hussein Obama II", and the answer of text question is "Joseph Robinette Biden Jr.", while the model erroneously interprets the response to the inquiry as the text displayed in the image. Secondly, the recognition of objects in images may not align with real-world knowledge. In video and 3D tasks, factual inaccuracies are influenced not only by object recognition errors but also by inconsistency across multiple perspectives. Furthermore, in audio comprehension tasks, the model is easily influenced by homophones, leading to factual errors.

\subsection{Conflict in Multimodal Fusion}
Many existing multimodal large models integrate information from different modalities using an adapter method \cite{liu2023visual}. The fusion at the hidden state can easily induce errors from the pre-training stage of different modalities. Hallucination generated from multiple modalities can influence each other.

In the pre-training stage of image-text pairing, prevailing methodologies predominantly leverage contrastive learning or an Encoder-Decoder framework to derive image embeddings. While these approaches have yielded commendable outcomes in conventional image classification tasks, they are not without limitations, notably in terms of occasional inaccuracies in image recognition and a propensity for hallucinations in capturing intricate details of images. For example, the question is: Are badminton rackets usually placed together with shuttlecocks? However, in the picture, the badminton racket is placed together with a basketball, and the model incorrectly answered the question as: No. These include residual hallucinations in responses to image-related prompts, attributable to a partial comprehension of the visual content and potential biases ingrained in the alignment process.

In the realm of audio-language models, the issue of "where is speaking part" emerges as a notable challenge, which can be categorized under this type of hallucination. This issue arises when the model inadvertently overlooks the audio feature preceding the target audio feature. For example, in Figure \ref{fig:example}, the content of the audio portion is: "Football plays a very important role in my life." while the entire model input is: "Here is a segment from an interview, [audio feature], what does the person being interviewed like?" The model occasionally confuse the features at the junction of text and audio during inference, which leads to the neglect or incorrect prediction of football in this example.

\section{Emergence for AGI Hallucination}

\subsection{Training Data Distribution}

The significance of training data on the efficacy of models is paramount. Both the quantity and quality of the data directly influence model's distribution \cite{hammoudeh2022training}. The generalization ability of the model such as overfitting and underfitting to specific data distributions can adversely impact model function, occasionally leading to hallucinations in the model's outputs. In the machine learning, \citet{Mougan2023ExplanationSH}'s research highlighted that shifts in training data distribution markedly alter explanation characteristics. This findings suggest that data distribution changes can profoundly influence a model's ultimate performance. \citet{Dong2023HowAI}'s work revealed that the quantity of compositional data significantly impacts model performance, whereas the ratio of composition bears minimal effect. Through integrating specialized domain data (mathematics and coding) into a general domain, the impact of varied data distributions on model performance was examined. The conclusion was that model capabilities are enhanced in scenarios with limited resources but diminish in high-resource contexts when compared to the capabilities derived from individual data sources. Furthermore, it was observed that as the model's size increases, there is a corresponding amplification in performance gains in low-resource settings, particularly for mathematical and general competencies. \citet{Dziri2022OnTO}'s findings indicate that the proportion of hallucinations in generated responses is substantially higher compared to that in the training data. LIMA \cite{Zhou2023LIMALI} posit that the vast majority of knowledge in large language models is acquired during the pretraining phase, and that only a minimal amount of instruction tuning data is requisite for training these models to generate outputs of high quality. \citet{Li2020OverfittingOU} stated that both overfitting and underfitting can impact the model's performance.

\subsection{Timeliness of Information}

Recent research underscores the critical importance of timeliness of information in preventing hallucinatory outputs from large models. Like humans, large models sometimes forget previous information. However, updating information in large models is quite challenging, leading to hallucinatory output due to outdated information. In the research, \citet{Lin2023SpecialityVG} has highlighted specific instances where foundational models, during the fine-tuning process, tend to sacrifice their general applicability in favor of becoming more specialized for particular tasks. Furthermore, \citet{Zhai2023InvestigatingTC}'s work on evaluating catastrophic forgetting in Multimodel Large Language Models (MLLMs) is notable. By treating each MLLM as an image classifier, Zhai discovered that initial fine-tuning stages on an image dataset not only enhance performance across various other image datasets but also improve the alignment between textual and visual features. However, as fine-tuning continues, a significant drawback emerges: the MLLMs start to hallucinate. This leads to a considerable decline in their generalizability, a trend that persists even when the image encoder component remains unchanged.

\subsection{Ambiguity in Different Modalities}
The hallucinations that occur between multiple modalities are attributed to interactions among these modalities. Ambiguities between different modalities can lead to the model outputting incorrect knowledge. For instance, in the case of images with text, a conflict between the textual content in the image and the knowledge in the accompanying text can cause the model to output incorrect knowledge from the image. Comparable issues are also observable in 3D and video modalities. \citet{Wang2023AnLM}'s research elucidates that hallucinations in LVLM primarily originate from discrepancies between visual and textual modalities.

\section{Mitigation for AGI Hallucination}

Due to the complexity of hallucinations and the black-box nature of neural networks, mitigating hallucinations is a difficult task. Current research in reducing hallucinations can be divided into different stages of the mitigation process, the data preparation (\textbf{Data}), the model training (\textbf{Train \& SFT \& RLHF}), and the model inference and post-processing (\textbf{Inference \& Post-hoc}). We will detail the work of these three parts as follows:

\begin{table}[t]
\centering
\scalebox{1}{
\begin{tabular}{lc}
\toprule
\rowcolor[gray]{.92} \textbf{Use-case} & \textbf{(\%) }  \\  \midrule
Generation & 45.6\% \\
Open QA &  12.4\% \\
Brainstorming & 11.2\% \\
Chat & 8.4\% \\
Rewrite & 6.6\% \\
Summarization & 4.2\% \\
Classification & 3.5\% \\
Other & 3.5\% \\
Closed QA & 2.6\% \\
Extract & 1.9\% \\
\bottomrule
\end{tabular}
}
\caption{The pre-training data distribution in \cite{Ouyang2022TrainingLM}.}
\label{tab:data}
\end{table}

\textbf{Data}\ \ \ 
In both the pre-training and fine-tuning stages, the distribution and quality of data are crucial. High-quality data usage during training can reduce model hallucinations. \citet{Ouyang2022TrainingLM}'s research shows that datasets with instructional annotations improve truthfulness compared to GPT-3 \cite{Brown2020LanguageMA}, signifying a major progression. They emphasized the importance of data distribution and quality in achieving robust model performance. Unlike previous approaches that handled each task individually, the data in Table \ref{tab:data} underscores the essential roles of tasks like Generation, OpenQA, and Brainstorming. Furthermore, the process of data cleansing is vital. For instance, the creators of Llama 2 \cite{Touvron2023Llama2O} intentionally upsample data from factual sources such as Wikipedia in constructing their pre-training corpus. Similarly, the developers of Falcon \cite{Penedo2023TheRD} expertly extract high-quality data from the web using heuristic rules, underscoring that well-curated training corpora are foundational for effective LLMs. \citet{Wang2023MitigatingFH}'s study highlights the importance of fine-grained captions, which align more precisely with image representations, thus reducing hallucinations in multimodal contexts. Additionally, improving the resolution of images in datasets is shown to further mitigate hallucinations induced by the model. \citet{Li2023VideoChatCV} proposed using detailed video descriptions to train a VideoChat model. Moreover, the introduction of high-quality 3D features (point clouds and textures) or audio features (phonemes and tones) can alleviate the hallucinatory manifestations of the model.

\begin{table*}[t!]
\centering
\scalebox{1}{
\begin{tabular}{ccp{8cm}}
\toprule
\rowcolor[gray]{.92} \textbf{Area} & \textbf{Stage} & \multicolumn{1}{c} {\textbf{Works}}  \\  
\midrule

\multirow{3}{*}{Language} & {\textit{Data}} & \citet{Brown2020LanguageMA}, \citet{Ouyang2022TrainingLM},  \citet{Touvron2023Llama2O}, \citet{Penedo2023TheRD} \\
\cmidrule{2-3}
& \textit{Train \& SFT \& RLHF} & \citet{Touvron2023LLaMAOA}, \citet{GPT4}, \citet{Rafailov2023DirectPO} \\
\cmidrule{2-3}
& \textit{Inference \& Post-hoc} & \citet{Krishna2023PostHE}, \citet{Cui2023ChatLawOL}, \citet{Feng2023KnowledgeST}, \citet{Melz2023EnhancingLI}, \citet{Zhao2021BoostingEI}, \citet{Shapkin2023EntityAugmentedCG} \\
\cmidrule{1-3}

\multirow{3}{*}{Visual Language} & \textit{Data} & \citet{Wang2023MitigatingFH} \\
\cmidrule{2-3}
& \textit{Train \& SFT \& RLHF} & \citet{liu2023visual}, \citet{Yoon2022InformationTheoreticTH}, \citet{Sun2023AligningLM} \\
\cmidrule{2-3}
& \textit{Inference \& Post-hoc} & \citet{Agarwal2021TowardsTU},  \citet{Peng2021MultimodalET}, \citet{Zhao2021BoostingEI},
\citet{Zhou2023AnalyzingAM}\\
\cmidrule{1-3}

\multirow{3}{*}{Video Language} & \textit{Data} & \citet{Li2023VideoChatCV} \\
\cmidrule{2-3}
& \textit{Train \& SFT \& RLHF} & \citet{li2023videochat}, \citet{jin2023chat}, \citet{yu2023deficiency}, \citet{zhang2023video}, \citet{shukor2023unified}, \citet{jin2023knowledge} \\
\cmidrule{2-3}
& \textit{Inference \& Post-hoc} & \citet{Jin2023KnowledgeConstrainedAG},\citet{yin2023woodpecker} \\
\cmidrule{1-3}

\multirow{2}{*}{Audio Language} & \textit{Train \& SFT \& RLHF} & \citet{Kanda2023FactualCO}, \citet{serai2022hallucination}, \citet{sridhar2023parameter},
\citet{doh2023lp}, \citet{Gong2023ListenTA}\\
\cmidrule{2-3}
& \textit{Inference \& Post-hoc} & \citet{lyu2023macaw} \\
\cmidrule{1-3}

\multirow{2}{*}{3D \& Agent} &  \textit{Train \& SFT \& RLHF} & \citet{Mu2022CertifiedPS}, \citet{li2023m3dbench}, \citet{gong2022posetriplet}, \citet{ren2023robots}, \citet{szot2023large}, \citet{xia2023kinematic} \\

\bottomrule
\end{tabular}
}
\caption{Overview of Migitation for AGI Hallucination.}
\label{tab:sumOfmitigation}
\end{table*}

\textbf{Train \& SFT \& RLHF}\ \ \ 
Numerous studies have demonstrated that applying suitable training techniques and loss functions during the training or supervise finetuning phase can significantly reduce model hallucination. LLaMA \cite{Touvron2023LLaMAOA} reveals that instruction fine-tuning leads to rapid enhancements in Massive Multitask Language Understanding (MMLU) performance. LLaVA \cite{liu2023visual} illustrates that a dual-stage training approach can effectively boost the efficacy of large multimodal models. In the first stage, the model synchronizes different modalities using image-caption datasets. Subsequently, in the second stage, the model is trained with multimodal instruction data. \citet{Yoon2022InformationTheoreticTH} developed the Text Hallucination Mitigating THAM framework, which integrates a THR (Text Hallucination Regularization) loss. This THR loss is derived from their novel information-theoretic approach to measuring text hallucination. \citet{Kanda2023FactualCO} have developed an optimization scheme for ASR (Automatic Speech Recognition) focused on enhancing factual consistency, with the aim of reducing hallucinations of the ASR model.

Recent advancements increasingly indicate that RLHF (Reinforcement Learning from Human Feedback) is an efficacious approach to mitigate hallucinations in machine learning. RLHF comprises four sub-modules: reference model, base model, reward model, and value model. Specifically, GPT-4 \cite{GPT4} employs RLHF to enhance alignment with human preferences, significantly reducing instances of model hallucinations. Recentely, DPO \cite{Rafailov2023DirectPO} (Direct Preference Optimization) , has been proposed as a novel RLHF strategy. This method leverages just two models to effectively discern between positive and negative samples which further curtailing hallucinations. \citet{Sun2023AligningLM} introduces a novel alignment algorithm named Factually Augmented RLHF. This innovative approach enhances the reward model by incorporating additional factual data, including image captions and verified multi-choice options. This augmentation significantly mitigates the issue of reward hacking commonly encountered in RLHF, thereby elevating the algorithm's overall efficacy. In the context of agent-based systems, RL (Reinforcement Learning) is pivotal during both the training and inference phases. Emerging studies \cite{Mu2022CertifiedPS} suggest that employing multi-agent RL training methodologies can substantially enhance the task completion efficacy of these agents. 

\textbf{Inference \& Post-hoc}\ \ \ Post hoc explanation methodologies in machine learning are primarily categorized into perturbation-based and gradient-based approaches \cite{Agarwal2021TowardsTU}. These approaches are designed to modify input features during the model's inference phase, aligning the model's output more closely with human interpretative preferences. Perturbation-based methods involve constructing an interpretable surrogate model from the original, more opaque model. This is achieved through systematic perturbations of the input samples. Conversely, gradient-based methods, exemplified by techniques such as SmoothGrad \cite{Smilkov2017SmoothGradRN} and Integrated Gradients, focus on computing the gradients of the model with respect to its input features. The process aids in determining the sensitivity of the model's output to each individual feature. In this context, \citet{Krishna2023PostHE} have developed an innovative technique In-Context Learning with Post Hoc Explanations. They automates the generation of rationales, effectively addressing the challenges presented by traditional post hoc explanation methods. LURE \cite{Zhou2023AnalyzingAM} uses the Revisor method to alleviate the LVLM hallucination by correcting the generated hallucination responses. \citet{Sridhar2023ParameterEA} introduce a parameter-efficient, inference-time, faithful decoding algorithm that facilitates the use of compact audio captioning models. The results demonstrate performance on par with larger counterparts trained on extensive datasets.

Leveraging a knowledge-base plays a crucial role in mitigating hallucinations during model inference and post-processing. ChatLaw \cite{Cui2023ChatLawOL} proposes a strategy to counteract hallucination by augmenting the training process of the model and integrating four distinct modules during inference, namely 'consult', 'reference', 'self-suggestion', and 'response'. \citet{Feng2023KnowledgeST} endeavor to instruct LLMs in querying relevant domain-specific knowledge from external knowledge graphs for answering specialized queries. \citet{Melz2023EnhancingLI} employs RAG (Retrieval Augmented Generation), \cite{Lewis2020RetrievalAugmentedGF}, to enhance problem-solving efficacy. The Verify-then-Edit methodology \cite{Zhao2021BoostingEI} is focused on improving the factuality of predictions through post-editing reasoning chains, using external information sourced from Wikipedia. \citet{Peng2021MultimodalET} adopts the information retrieval paradigm, automating the identification of related entities in text-image pairs to boost model performance. \citet{Zhao2021BoostingEI} introduces an innovative approach that constructs a MMKG (multi-modal knowledge graph), linking visual objects with named entities and simultaneously capturing the relationships among these entities, aided by external knowledge obtained from the web. \citet{Jin2023KnowledgeConstrainedAG} utilizes ClipBERT for extracting video-question features and derives frame-wise object-level external knowledge from a commonsense database to augment the performance of the model. \citet{Shapkin2023EntityAugmentedCG} introduces an end-to-end trainable architecture that incorporates a scalable entity retriever directly into the LLM decoder. This method significantly enhances the performance of code generation.

\begin{table*}[t!]
\centering
\scalebox{1}{
\begin{tabular}{ccp{8cm}}
\toprule
\rowcolor[gray]{.92} \textbf{Area} & \textbf{Eval-Mod} & \multicolumn{1}{c} {\textbf{Works}}  \\  
\midrule

\multirow{4}{*}{Language} & {\textit{Benchmark}} & \citet{Friel2023ChainpollAH}, \citet{Bang2023AMM}, \citet{Yu2023KoLACB}, \citet{Liang2023UHGEvalBT}, \citet{Ramakrishna2023INVITEAT},\citet{Sun2023HeadtoTailHK},\citet{Sadat2023DelucionQADH} \\
\cmidrule{2-3}
& \textit{Rule-Based} & \citet{Lin2004ROUGEAP}, \citet{Papineni2002BleuAM}, \citet{Sun2023HeadtoTailHK}, \citet{Min2023FActScoreFA}, \citet{Yu2023KoLACB} \\
\cmidrule{2-3}
& \textit{Large Model-Based} & \citet{Zhang2019BERTScoreET}, \citet{Yuan2021BARTScoreEG}, \citet{Zha2023AlignScoreEF}, \citet{Chan2023CLAIREI},\citet{Sarmah2023TowardsRH}, \citet{Chan2023ChatEvalTB} \\
\cmidrule{2-3}
& \textit{Human-Based} & \citet{Chiang2023CanLL}, \citet{Min2023FActScoreFA} \\
\cmidrule{1-3}

\multirow{3}{*}{Visual Language} & \textit{Benchmark} & \citet{Bang2023AMM}, \citet{Liu2023MMBenchIY}, \citet{Yu2023MMVetEL}, \citet{Li2023SEEDBenchBM}, \citet{Xu2023LVLMeHubAC},\citet{Villa2023BehindTM}, \citet{Gunjal2023DetectingAP}, \citet{Liu2023HallusionBenchYS}, \citet{Hu2023RSGPTAR} \\
\cmidrule{2-3}
& \textit{Rule-Based} & \citet{Shukor2023BeyondTP}, \citet{Vedantam2014CIDErCI}, \citet{Rohrbach2018ObjectHI}, \citet{Li2023EvaluatingOH}, \citet{Lovenia2023NegativeOP} \\
\cmidrule{2-3}
& \textit{Large Model-Based} & \citet{Zhai2023HallESwitchCO}, \citet{Bai2023TouchStoneEV} \\
\cmidrule{2-3}
& \textit{Human-Based} & \citet{guan2023hallusionbench}, \citet{Xu2023LVLMeHubAC}, \citet{Yin2023LAMMLM} \\
\cmidrule{1-3}

\multirow{3}{*}{Video Language} & \textit{Benchmark} & \citet{Bang2023AMM}, \citet{Chen2023AutoEvalVideoAA}, \citet{Chen2023AutoEvalVideoAA}, \citet{Mangalam2023EgoSchemaAD}, \citet{Liu2023EvalCrafterBA}, \citet{Li2023MVBenchAC}, \cite{feng2023llm4vg}, \citet{Himakunthala2023LetsTF}, \citet{Liu2023ModelsSH}, \citet{grauman2023ego} \\
\cmidrule{2-3}
& \textit{Rule-Based} & \citet{Vedantam2014CIDErCI}, \citet{Rohrbach2018ObjectHI}  \\
\cmidrule{2-3}
& \textit{Large Model-Based} & \citet{zhang2023simple}, \citet{feng2023llm4vg}, \citet{chen2023autoeval} \\
\cmidrule{2-3}
& \textit{Human-Based} & \citet{ma2023large}, \citet{chen2023egoplan} \\
\cmidrule{1-3}

\multirow{2}{*}{Audio Language} & \textit{Benchmark} & \citet{behera2023aquallm}, \citet{de2023emphassess} \\
\cmidrule{2-3}
& \textit{Rule-Based} & \citet{yu2023hourglass} \\
\cmidrule{2-3}
\cmidrule{1-3}

\multirow{2}{*}{3D \& Agent} &  \textit{Benchmark} & \citet{behera2023aquallm}, \citet{chen2023egoplan} \\
\cmidrule{2-3}
& \textit{Large Model-Based} & \citet{yang2023llm} \\
\cmidrule{2-3}
& \textit{Human-Based} & \citet{Yin2023LAMMLM} \\

\bottomrule
\end{tabular}
}
\caption{Overview of Evaluation for AGI Hallucination.}
\label{tab:sumOfevaluation}
\end{table*}

\section{Evaluation for AGI Hallucination}

The evaluation of AGI hallucinations is critically important, with substantial progress being made in AGI assessment in current research. Human evaluation of model-generated hallucinations is a time-consuming process, also humans themselves are susceptible to hallucinations. In the field of LLMs, \citet{Yu2023KoLACB, Sun2023HeadtoTailHK} have proposed the use of rule-based methods to detect model hallucinations, while \citet{Zha2023AlignScoreEF, Min2023FActScoreFA, Liu2023ModelsSH,Dhuliawala2023ChainofVerificationRH} have introduced automated assessment methods for detecting factual hallucinations, both of which are commendable contributions in automatically evaluate large model hallucinations. The occurrence of hallucinations in multimodal AGI is even more complex, encompassing not only misconceptions arising from language comprehension errors but also hallucinations related to the recognition of information in other modalities. In the context of LVLMs, POPE \cite{Li2023EvaluatingOH} first proposed the use of discriminative methods to detect object hallucinations in images. Additionally, the work of \cite{Zhang2019BERTScoreET, Yuan2021BARTScoreEG, Bai2023TouchStoneEV} highlights that methods based on LLMs can more effectively detect hallucinations. External knowledge bases also play a vital role in hallucination detection. In this section, we first introduce the benchmark for hallucination evaluation, 
 then we categorize hallucination assessment into three types: rule-based, large model-based, and human-based.
 
\textbf{Benchmark}\ \ \ In both the traditional deep learning era and the era of large models, benchmarks play a crucial role in evaluating model illusions. Benchmarks in different fields, such as ChainPoll \cite{Friel2023ChainpollAH} and \citet{Bang2023AMM}, focus on assessing specific domain weaknesses, while benchmarks across different modalities evaluate various model skills. KoLA \cite{Yu2023KoLACB} introduced a knowledge-driven benchmark comprising 19 tasks, utilizing Wikipedia data along with continuously collected emerging corpora to construct more granular evaluations, addressing unseen data and evolving knowledge. UHGEval \cite{Liang2023UHGEvalBT} is an Unconstrained Hallucination Generation Evaluation benchmark for evaluating prominent Chinese language models. INVITE \cite{Ramakrishna2023INVITEAT} automatically generates invalid questions to assess large model illusions. Meanwhile, Head-to-Tail \cite{Sun2023HeadtoTailHK} uses a template-based method to automatically generate 18K QA (question-answer) pairs for model illusion assessment. DELUCIONQA \cite{Sadat2023DelucionQADH} employs a QA system to automatically generate domain-specific questions and answers for detecting model illusions. Numerous benchmarks have also emerged in the fields of image and video, particularly for LVLMs such as MMBench \cite{Liu2023MMBenchIY}, MM-Vet \cite{Yu2023MMVetEL}, SEED-Bench \cite{Li2023SEEDBenchBM}, LVLM-eHub \cite{Xu2023LVLMeHubAC}, and AutoEval-Video \cite{Chen2023AutoEvalVideoAA}. MERLIM \cite{Villa2023BehindTM} is an extensive database with over 279,000 image-question pairs, primarily focusing on identifying and analyzing cross-modal 'hallucination' events in IT-LVLMs (Image-Text Language-Vision Language Models). In multimodal contexts, finer-grained captions can detect more nuanced hallucinations in image-text interactions, \citet{Gunjal2023DetectingAP} with a dataset of 16k finely annotated VQA examples. HALLUSIONBENCH \cite{Liu2023HallusionBenchYS} is the first to consider visual illusion and knowledge hallucination of LVLMs. RSGPT \cite{Hu2023RSGPTAR} introduces a benchmark concerning Remote Sensing. In the video domain, benchmarks like AutoEval-Video \cite{Chen2023AutoEvalVideoAA} , EgoSchema \cite{Mangalam2023EgoSchemaAD}, EvalCrafter \cite{Liu2023EvalCrafterBA}, and MVBench \cite{Li2023MVBenchAC} have emerged. AutoEval-Video and MVBench establish video QA datasets, assessing Multi-modal Video from multiple perspectives. EgoSchema focuses on Very Long-form Video Language Understanding, while EvalCrafter and LLM4VG \cite{feng2023llm4vg} emphasizes the assessment of Video Generation. \citet{Himakunthala2023LetsTF} evaluates complex video reasoning tasks, and \citet{Liu2023ModelsSH} evaluates the factuality datasets for video captioning. \citet{grauman2023ego} present an exocentric video of skilled human activities as multimodal multiview video dataset and benchmark challenge. \citet{behera2023aquallm} used LLM to generate expansive, high-quality Audio Question Answering datasets, contributing significantly to the progression
of Audio research.

\textbf{Rule-Based}\ \ \ Rule-based evaluation methodologies have significantly evolved in traditional deep learning tasks, exemplified by the development of ROUGE \cite{Lin2004ROUGEAP} and BLEU \cite{Papineni2002BleuAM}. However, the challenge intensifies when assessing hallucinations in single sentences, particularly in the context of LLMs. In the evaluation of LLMs,  Head-to-Tail \cite{Sun2023HeadtoTailHK} , employing a triad of metrics: accuracy (A), hallucination rate (H), and missing rate (M), to gauge the quality of the generated sentences. This approach further integrates exact match (EM), token F1 (F1), and ROUGE-L into its assessment criteria. \citet{Min2023FActScoreFA}, firstly, breaking a generation into a series of atomic facts—short statements, then assigning a binary label to each atomic fact, and finally evaluating the FACTSCORE in the sentence.

Furthermore, KoLA \cite{Yu2023KoLACB} initiative proposed the innovative Self-contrast Metric, rooted in ROUGE-L. This metric computes an average contrast score by juxtaposing results derived from knowledge-based prompts, standard prompts, and human responses, thereby effectively evaluating hallucinations in LLM-generated content.

For LVLMs, \citet{Shukor2023BeyondTP} adopted a unique approach, utilizing VQA accuracy, CIDEr \cite{Vedantam2014CIDErCI}, and CHAIRs \cite{Rohrbach2018ObjectHI}, specifically for appraising hallucinations within the COCO dataset. Object Hallucination, in particular, represents a formidable challenge in LVLMs. To address this problem, POPE \cite{Li2023EvaluatingOH} , based on the co-occurrence principle of objects in images. POPE, a polling-based query methodology, is designed to scrutinize model hallucinations on objects through binary "is or not" questions. In a parallel development, NOPE \cite{Lovenia2023NegativeOP} leveraging LLMs to generate data featuring NegP (Negative Pronouns). NOPE's approach is instrumental in evaluating object hallucination in these models.

\textbf{Large Model-Based}\ \ \ 
The surge in popularity of Large Model-Based evaluation techniques is quite remarkable, particularly in contrast to Rule-Based methods. These Model-Based approaches excel in executing detailed assessments of models. BERTSCORE \cite{Zhang2019BERTScoreET} employs the BERT model to evaluate the similarity between candidate and reference sentences.  BARTSCORE \cite{Yuan2021BARTScoreEG} assesses model-generated sentences by calculating weighted marginal probability distributions, utilizing token probability distributions at each timestep. ALIGNSCORE \cite{Zha2023AlignScoreEF} establishes a novel text-to-text information alignment function for assessing the factual consistency between two text pieces. CLAIR \cite{Chan2023CLAIREI} leverages the capabilities of highly credible LLMs like ChatGPT to generate both candidate scores and an explanatory rationale for the score. In the financial reporting sector, \citet{Sarmah2023TowardsRH} integrates BERTScore, BARTScore, Jaro similarity, and the Longest Common Subsequence method for evaluating model-generated reports. ChatEval \cite{Chan2023ChatEvalTB} employs a team of multi-agent referees to assess the response quality produced by various models in response to open-ended inquiries and conventional tasks in the domain of NLG (Natural Language Generation). HALLE-SWITCH \cite{Zhai2023HallESwitchCO} proposes CCEval, an innovative GPT-4 assisted evaluation tool for detailed captioning. Finally, TouchStone \cite{Bai2023TouchStoneEV} employs GPT-4, a formidable language model, to assess the diverse capabilities of LVLMs. \citet{Kabra2023EvaluatingVF} conducted an evaluation of VLMs for 3D Objects, focusing on their responses to variations in object view, the phrasing of questions, previously inferred information specified in the prompt, and the accessibility of the object's visual characteristics.

\textbf{Human-Based}\ \ \ Human-based evaluation in assessing the precision and authenticity of models, particularly in aligning with human preferences. \citet{Chiang2023CanLL} emphasizes the use of both human evaluation and LLM evaluation in two specific NLP tasks: open-ended story generation and adversarial attacks. The findings indicate that evaluations conducted by LLMs can lead to ethical concerns.

\citet{Min2023FActScoreFA} involves human annotators assigning one of three labels (Irrelevant, Supported, Not-supported) to each atomic fact, enhancing the precision of the assessment. In contrast, \citet{guan2023hallusionbench} entails human experts manually collecting 346 images across a variety of topics and types. These experts not only select each image meticulously but also compose corresponding question-answer pairs.

\citet{Liu2023ModelsSH} has developed an annotation protocol aimed at guiding annotators in evaluating and labeling the factuality of video captions. This method led to the creation of two human-annotated factuality datasets: ActivityNet-Fact, comprising 200 videos and 3,152 sentences, and YouCook2-Fact, including 100 videos and 3,400 sentences.

Additionally, the passage mentions LVLM-eHub \cite{Xu2023LVLMeHubAC} and LAMM \cite{Yin2023LAMMLM}, which utilize existing public datasets from various computer vision tasks for evaluation purposes. These evaluations are conducted either by human annotators or GPT models. LAMM, in particular, extends its scope to encompass a broad range of vision tasks in both 2D and 3D domains.

\section{Discourse for AGI Hallucination}
Mitigating hallucinations is essential in AGI models, it is also important to notice that not all such occurrences are detrimental.  In some scenarios, hallucinations can induce the model's creativity. Striking a balance between hallucination and creation is a crucial challenge.

Recent work has also shown that hallucinations are not entirely erroneous. Hallucinations play a role as adversarial examples to increase the robustness and creation of models, which need to produce hallucinations in contextually reasonable situations - akin to 'white lie' in human life. \citet{Yao2023LLMLH}'s research utilized weak semantic prompts and OOD prompts to elicit hallucinatory responses from LLMs. This approach led to a reassessment of hallucinations as a different perspective on adversarial examples. \citet{Zhang2023UserControlledKF} focused on the degree of faithfulness to reference knowledge in generated responses, striving for a balance between creativity and hallucination. \citet{Zhao2023BeyondHE} employed a method called Hallucination-Aware Direct Preference Optimization, using hallucinatory samples to optimize models through reinforcement learning, demonstrating this method's effectiveness in mitigating hallucinations in LVLMs. \citet{Qiu2022IterativeTB} introduced an approach where a teacher iteratively generates synthetic training data based on the learner's status, a process termed data hallucination teaching. \citet{Wu2023HallucinationIT} developed a 'Hallucinator' to generate additional positive image samples for enhanced contrast training. \citet{Fei2023SceneGA} conceived a visual scene hallucination mechanism that dynamically creates pseudo visual scene graphs from textual scene graphs, significantly improving inference-time image-free unsupervised multimodal machine translation. \citet{Kulal2023PuttingPI} proposed inserting characters into scenes, enabling models to generate videos with both character and scene hallucinations, achieving harmonious composition and creativity. \citet{McKee2021MultiObjectTW} initially added video simulation augmentations to create hallucinated video data and then trained a tracker jointly on this hallucinated data and mined hard video examples. HALLUAUDIO \cite{Yu2023HalluaudioHF} leverages a special audio format by hallucinating high-frequency and low-frequency parts as structured concepts for few-shot audio classification. Finally, \citet{Shah2023HaLPHL} pioneered the use of hallucinated latent positives in a skeleton-based CL (Contrastive Learning) framework. They also employed a MoCo-based framework that mixes latent space features of positives and negatives, better utilizing hallucinations in action generation tasks.

\begin{figure}
    \centering
    \resizebox{.49\textwidth}{!}{
    \includegraphics{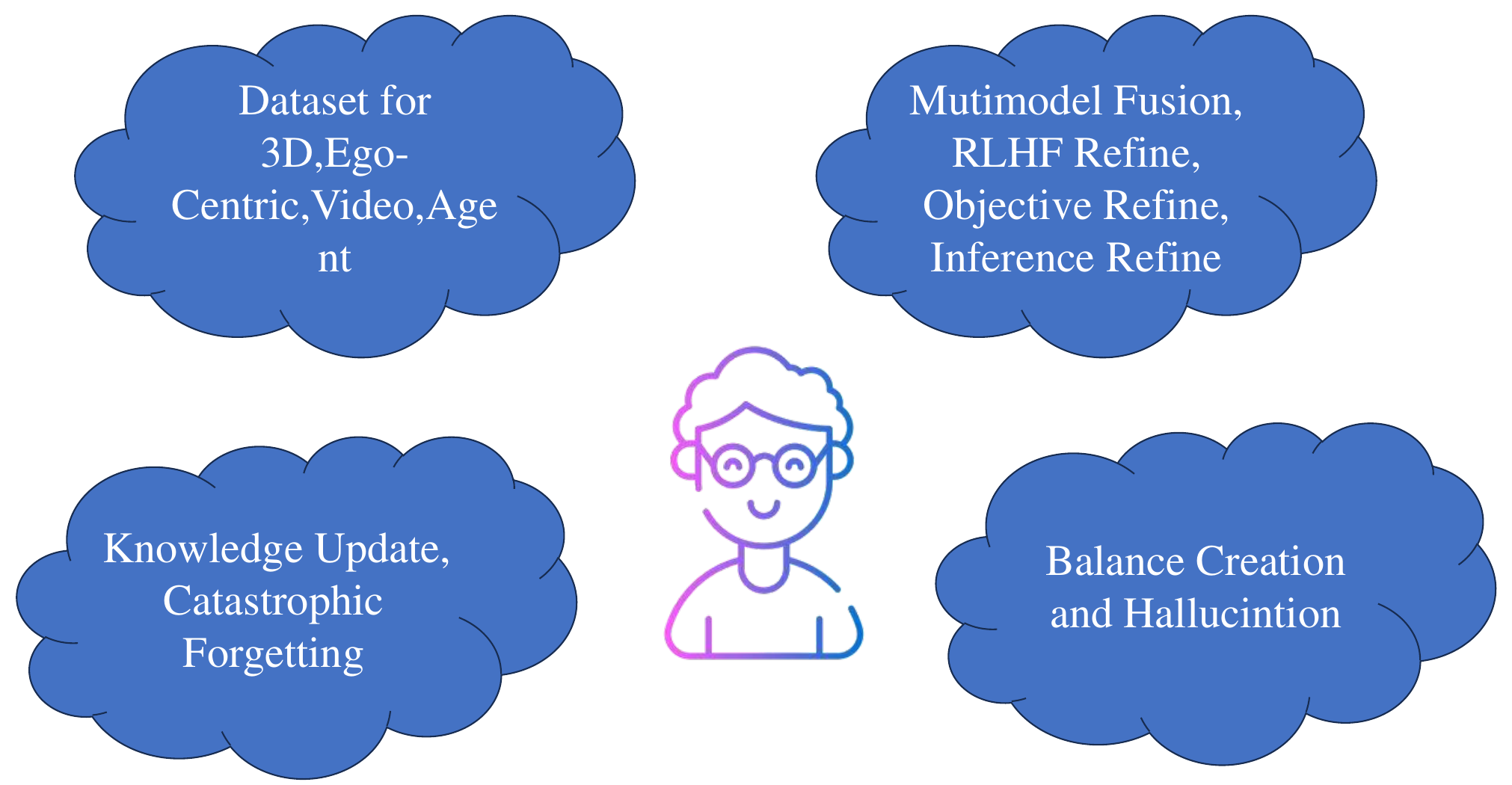}}
    \caption{
    Talking about Future.}
    \label{fig:Future}
\end{figure}
\section{Talking about Future}
The model generate hallucinations is influenced by data, model architectures, and information updating. Figure \ref{fig:Future} shows the future about AGI (Hallucination) Research. Presently, there's a notable deficit of high-quality data in the spheres of audio, 3D modeling, and agent-based systems. Future studies must prioritize the development of robust datasets in these areas. It is notice whether tokens are suitable representations for modalities that non-textual context and how models across different modalities can be better integrated. For knowledge updates, it's imperative for models to emulate human-like short-term and long-term memory functions. Investigating methods to enhance knowledge updating in models while retaining their foundational information is a critical area of research. Additionally, the equilibrium between hallucination and creation is crucial. Investigating methodologies to enable models to appropriately engage in hallucinatory activities presents a significant and interesting research avenue.

\textit{Dispel the clouds of hallucination around the AGI, and build a true AGI.}

\bibliography{latex/acl_latex}

\appendix

\section{Example Appendix}
\label{sec:appendix}

\begin{figure*}
    \centering
    \resizebox{0.89\textwidth}{!}{
    \includegraphics{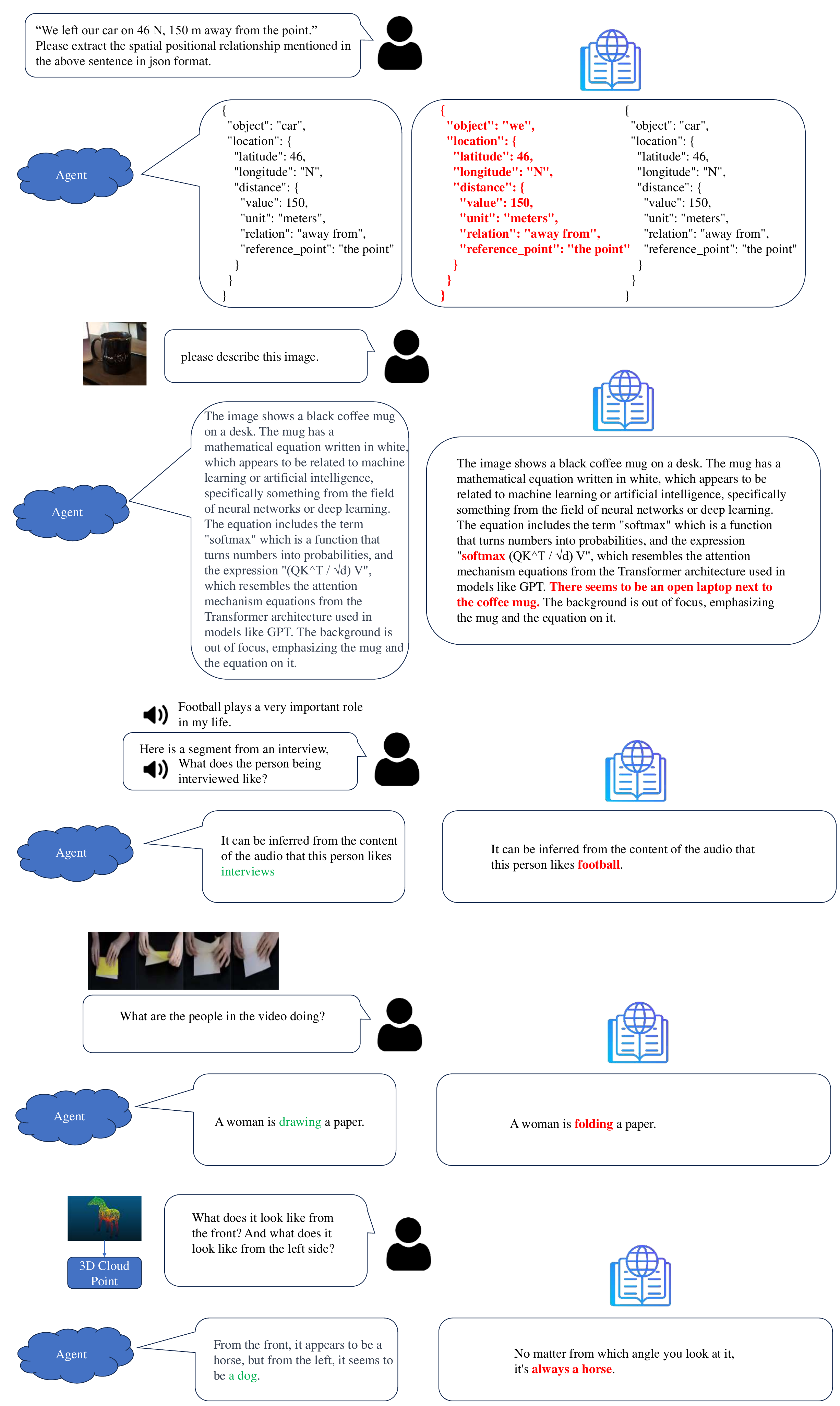}}
    \caption{
    Conflict in Intrinsic Knowledge examples.}
    \label{fig:example}
\end{figure*}

\begin{figure*}
    \centering
    \resizebox{1\textwidth}{!}{\includegraphics{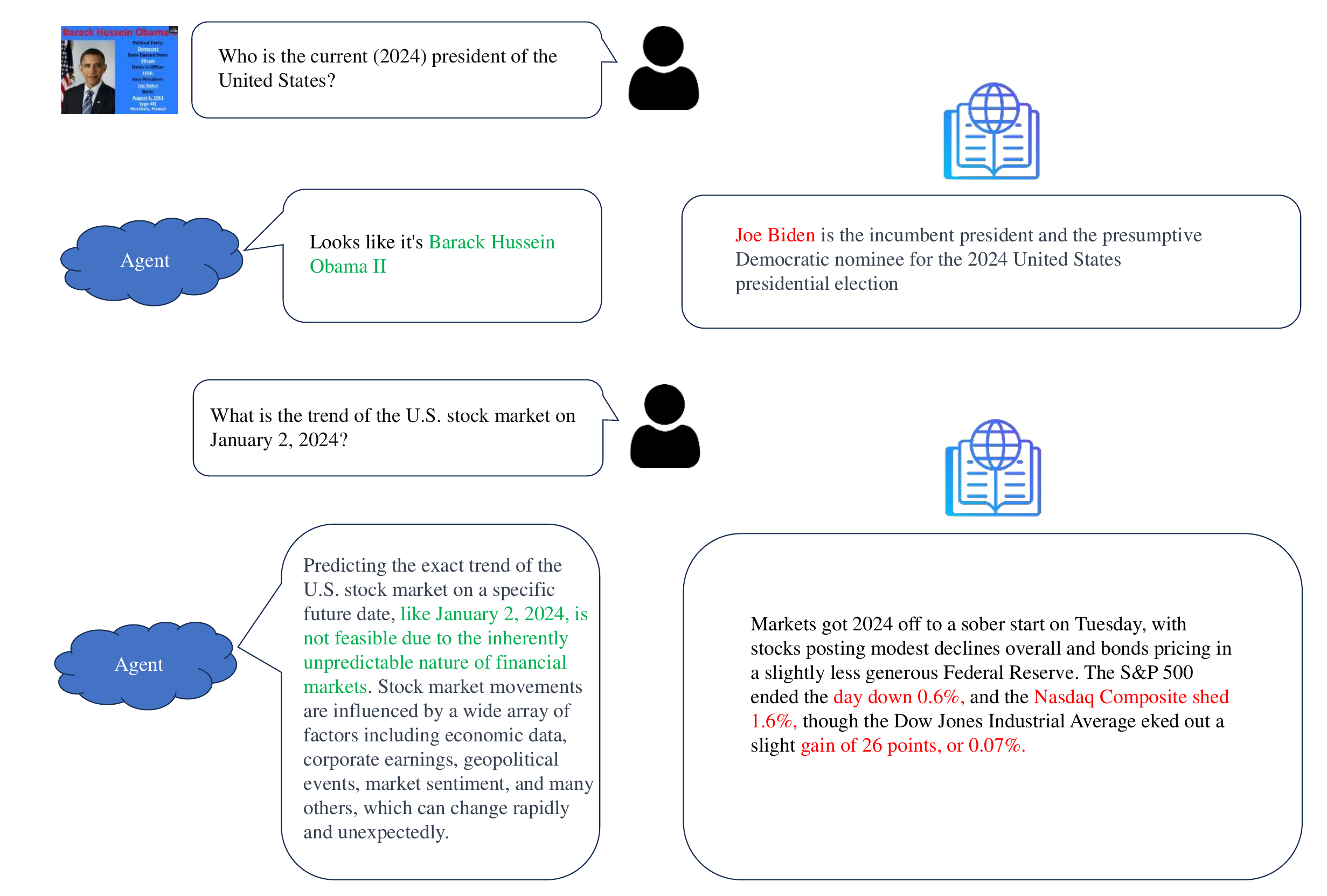}}
    \caption{
    Factual Conflict examples.}
    \label{fig:fact_example}
\end{figure*}

\end{document}